\documentclass{article}

\usepackage[preprint]{corl_2026} 

\usepackage{adjustbox}
\usepackage[ruled, vlined]{algorithm2e}
\usepackage{amsfonts}
\usepackage{amsmath}
\usepackage{amssymb}
\usepackage{amsthm}
\usepackage{array}  
\usepackage{booktabs}
\usepackage{capt-of}
\usepackage[font=small,skip=2pt]{caption}
\usepackage{subcaption}
\usepackage{colortbl}
\hypersetup{
    pagebackref=true,
    breaklinks=true,
    colorlinks=true,
    bookmarks=false
}

\usepackage{inconsolata}  
\usepackage{graphicx}
\usepackage{listings}
\usepackage{makecell}
\usepackage{multicol}
\usepackage{multirow}
\usepackage[numbers]{natbib}
\usepackage{nowidow}  
\usepackage{paralist}
\usepackage{pifont}
\usepackage{times}
\usepackage{tabularx}
\usepackage{thmtools}  
\usepackage{wrapfig}
\usepackage{xcolor}
\usepackage{svg}

\definecolor{flodarkpurple}{rgb}{0.288,0.1196,0.7}

\definecolor{ganeshamber}{rgb}{1.0, 0.75, 0.0}

\newcommand{\xmark}{\ding{55}}%

\usepackage{xcolor}
\usepackage{pifont}
\usepackage{booktabs}
\usepackage{colortbl}
\usepackage{makecell}
\usepackage{float}
\usepackage{graphicx} 

\usepackage{caption}
\usepackage{enumitem}

\usepackage{titlesec}

\newcommand{\coolname}{\textsc{DreamSteer}}

\title{\coolname: Latent World Models can steer VLA Policies during deployment without any finetuning}

\author{
  \textbf{Hanchen Cui}$^{1,2,*}$, 
  \textbf{Sergio Arnaud}$^{1}$, 
  \textbf{Arjun Majumdar}$^{1}$, 
  \textbf{Daniel Dugas}$^{1}$, 
  \textbf{Elie Aljalbout}$^{1}$, \\
  \textbf{Karthik Desingh}$^{2\dagger}$, 
  \textbf{Krishna Murthy Jatavallabhula}$^{1\dagger}$, 
  \textbf{Franziska Meier}$^{1\dagger}$ \\[2mm]
  $^{1}$Fundamental AI Research (FAIR), Meta \qquad
  $^{2}$University of Minnesota Twin Cities \\
  \vbox{\small\vspace{2mm}
    \centerline{$^{*}$Work done during an internship at Meta \qquad $^{\dagger}$Joint last authors}
  }
}

\begin{document}
\maketitle

\begin{abstract}
Pretrained vision--language--action (VLA) policies show promising zero-shot generalization, but often fail under deployment-time distribution shift, leading to decreased robustness and inconsistent instruction following. While prior work commonly tackles this by finetuning on in-distribution data, it assumes demonstrations collected on tasks in the target environment. In this work, we propose \textbf{\coolname}, a deployment-time steering framework for pretrained VLAs \textbf{without any finetuning or parameter modifications}. The key insight in \coolname{} is to leverage a \emph{latent world model} and a \emph{value model} to \emph{steer} pretrained VLA policies. During deployment, \coolname{} samples candidate action chunks from a VLA policy and predefined motion primitives, imagines their outcomes using an action-conditioned latent world model, and ranks the imagined trajectories with a language-conditioned value model. Across four real-world manipulation benchmarks with unseen objects, \coolname{} improves task success rate from 23.75\% to 66.25\% and instruction-following accuracy from 38.75\% to 56.25\% over the base VLA policy. Videos are available on the project website: https://dream-steer.github.io/.
\vspace{-0.5em}
\end{abstract}

\vspace{-0.3em}

\begin{figure}[h]
    \centering
    \includegraphics[width=\textwidth]{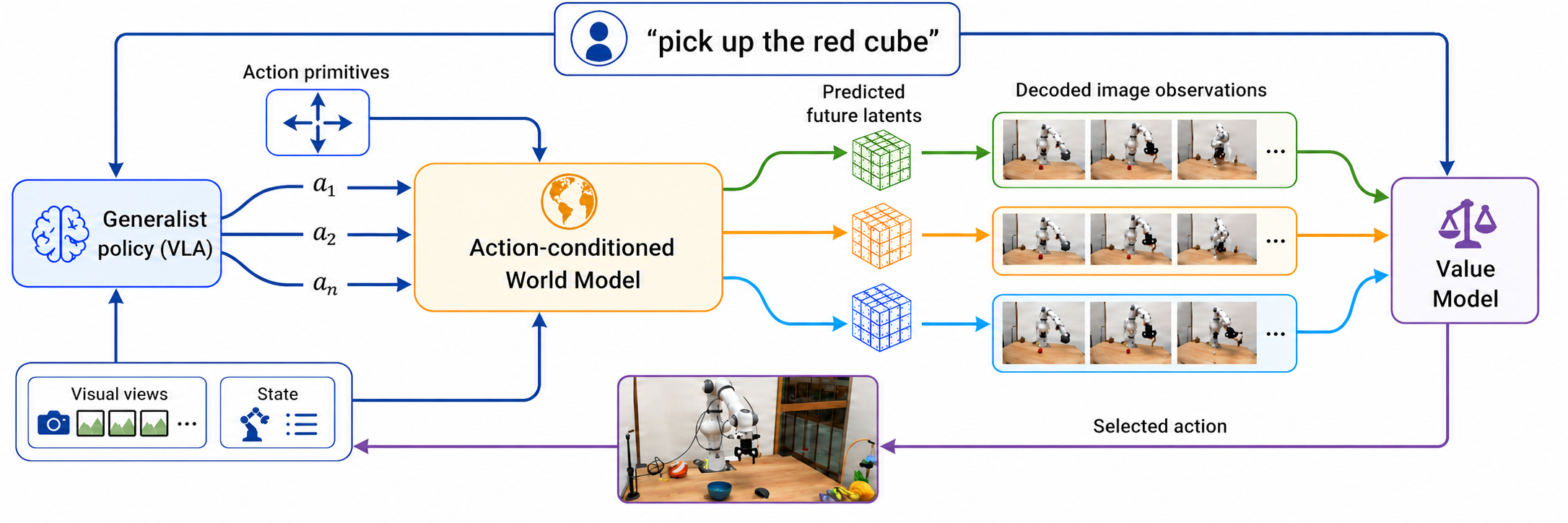}
    \caption{\textbf{\coolname: deployment-time policy steering.}
    A frozen VLA policy proposes candidate action chunks, which are augmented with a small set of predefined Cartesian action primitives. A latent world model predicts future observations, and a language-conditioned value model ranks the resulting trajectories before execution. \textbf{All models are trained frozen, and no target-environment data is used during model training}.}
    \label{fig:dreamsteer}
\end{figure}


\section{Introduction}

Recent progress in large-scale multimodal pretraining has enabled VLA policies~\citep{black2410pi0,bjorck2025gr00t,kim2024openvla,zitkovich2023rt,pertsch2025fast} that integrate perception, language understanding, and control within a unified framework. Trained on large-scale robot datasets, these models show promising transfer to new tasks, objects, and environments. However, pretrained VLAs remain brittle under deployment-time distribution shift, often failing on unseen objects or producing behaviors that violate language instructions. While finetuning can improve performance, it requires additional target-domain data and modifies the pretrained policy, which may not be desired or feasible. This raises a key question: \textbf{how can we improve pretrained VLA policies at deployment time without target-domain data finetuning?}

In the context of large language models, a similar challenge is tackled by \emph{test-time steering} approaches that sample candidate outputs and rerank them using external judges or scoring models to improve reliability without retraining~\citep{kang2025scalable,pi2025mr}. 
Pretrained VLAs provide a natural analogue: diffusion-based policies~\citep{black2410pi0,bjorck2025gr00t} and autoregressive VLAs~\citep{kim2024openvla,zitkovich2023rt,pertsch2025fast} both produce stochastic action samples that can be steered at inference time. However, unlike text outputs, the quality of an action chunk cannot be evaluated directly before execution in the physical world.

World models~\citep{agarwal2025cosmos,assran2025v,zhou2411dino} provide a natural mechanism for evaluating candidate actions before execution. Conceptually, they enable \emph{thinking before acting}. Given the current observation and a candidate action chunk, an action-conditioned world model can predict the resulting future observations. These imagined rollouts can then be ranked by a language-conditioned value model, enabling look-ahead action evaluation at deployment time.

\begin{wrapfigure}{r}{0.55\textwidth}
    \centering
    \includegraphics[width=\linewidth]{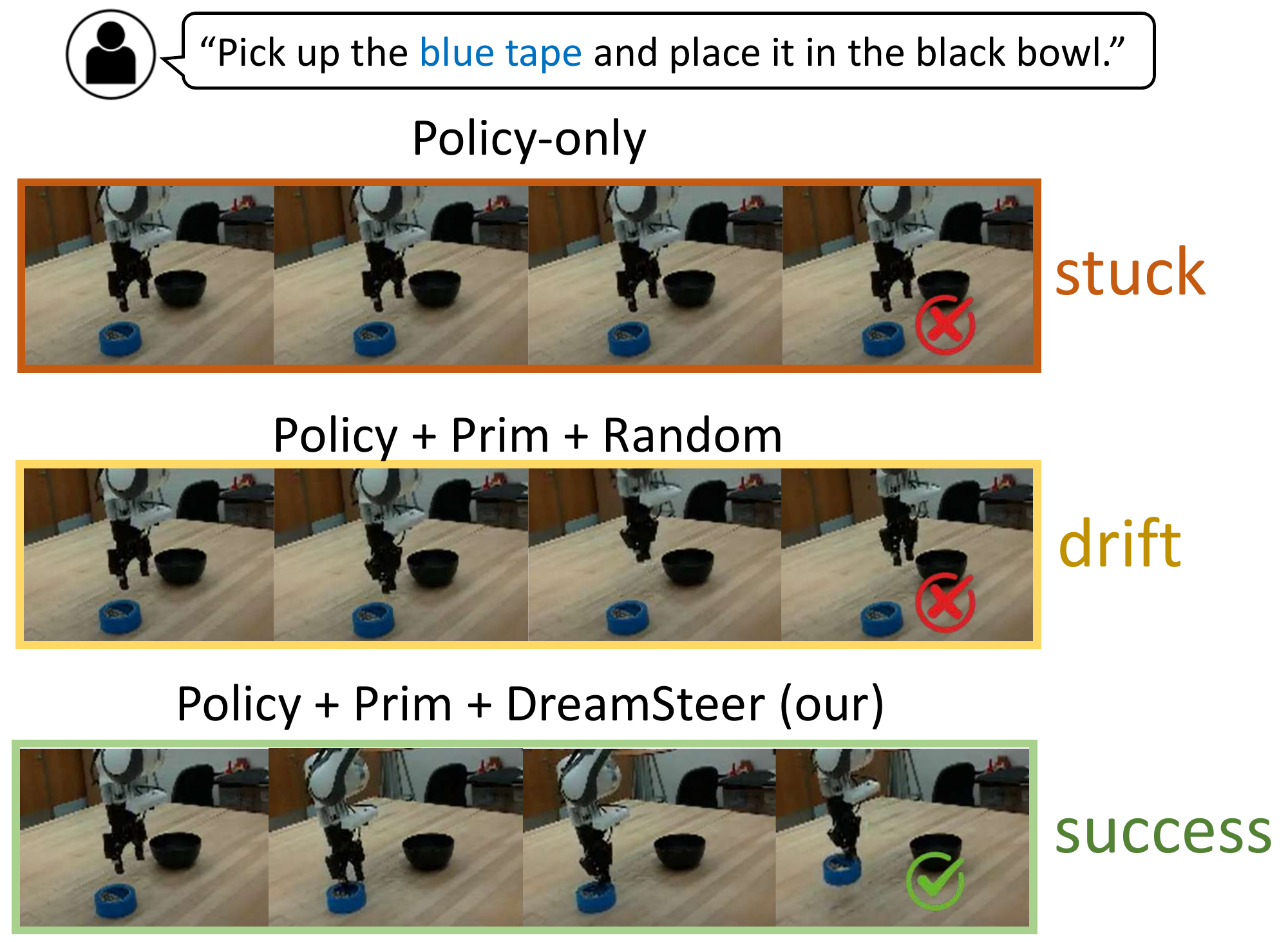}
    \caption{\textbf{\coolname{} improves deployment-time robustness} of a pretrained, frozen VLA policy.}
    \label{fig:3rollout}
\end{wrapfigure}

In this paper, we introduce \textbf{\coolname{}}, a deployment-time steering framework for pretrained VLA policies using latent world models with zero finetuning. Given a language instruction and current observation, \coolname{} samples multiple candidate action chunks from a pretrained VLA policy and augments them with predefined motion primitives. An action-conditioned latent world model predicts imagined future observations for each candidate, and a language-conditioned value model ranks the resulting rollouts. \coolname{} then executes the highest-scoring action chunk in the real environment. \textbf{All components remain fixed during deployment, and no target-environment data is used for adaptation}. As illustrated in Fig.~\ref{fig:3rollout}, single-sample policy execution may fail under deployment-time distribution shift, while evaluating multiple imagined futures before execution allows \coolname{} to select more reliable actions.

Our contributions are summarized as follows:
\begin{enumerate}[leftmargin=*, noitemsep, topsep=2pt]
\item We introduce \textbf{\coolname{}}, a deployment-time steering framework that evaluates candidate VLA action chunks through imagined latent-world-model rollouts before execution.

\item \coolname{} composes a pretrained VLA policy, a generalized latent world model, and a language-conditioned value model at deployment time, without finetuning any component on target-environment data.

\item We demonstrate \coolname{} improves OOD object manipulation success from 23.75\% to 66.25\%, and instruction-following accuracy from 38.75\% to 56.25\% over the $\pi_0$ VLA policy.
\end{enumerate}

\providecommand{\cmark}{}
\providecommand{\xmark}{}
\renewcommand{\cmark}{\ding{51}}
\renewcommand{\xmark}{\ding{55}}
\definecolor{rowgray}{gray}{0.92}

\section{Related Work}

\textbf{Generalist visuomotor policies.}
Large-scale multimodal pretraining has enabled vision--language--action (VLA) models to emerge as generalist manipulation policies. Recent models such as $\pi_0$~\citep{black2410pi0}, GR00T~\citep{bjorck2025gr00t}, OpenVLA~\citep{kim2024openvla}, RT-2~\citep{zitkovich2023rt}, and FAST~\citep{pertsch2025fast}, trained on large-scale datasets including DROID~\citep{khazatsky2024droid}, exhibit a broad range of manipulation
behaviors. Despite their broad capabilities, pretrained VLAs often remain brittle under deployment-time distribution shift~\citep{nakamoto2024steering,wu2025foresight,qi2025strengthening}, producing semantically incorrect behaviors or failing on unseen objects and environments. Instead of collecting new data and finetuning the policy, \coolname{} treats the pretrained VLA as a fixed generative action prior and steers sampled action chunks at deployment time.

\textbf{Robot World Models.}
World models predict future states conditioned on observations and actions, enabling agents to evaluate action consequences before execution~\citep{agarwal2025cosmos,assran2025v,zhou2411dino}. In robotics, world models support planning, policy evaluation, and data generation, with approaches differing primarily in prediction space and action conditioning. Pixel-space models such as Cosmos~\citep{agarwal2025cosmos} and DreamGen~\citep{jang2025dreamgen} generate visually rich futures but remain computationally expensive for online steering. In contrast, latent models such as DINO-WM~\citep{zhou2411dino} and VT-WM~\cite{higuera2026visuo} directly predict future states in compact latent space conditioned on robot actions, making them better suited for inference-time action evaluation.
\coolname{} uses an action-conditioned latent world model for efficient rollout and candidate evaluation. Candidate action chunks are rolled out in latent space, then decoded into image observations. This avoids expensive pixel-space video generation while preserving an image-based interface for evaluation.

\textbf{Policy Steering and Model-Based Policy Refinement.}
This class of methods improves pretrained policies using action proposals, predictive models, and external evaluators. Most policy steering approaches follow this decomposition: a policy proposes actions, a world model rolls out their consequences, and a value or reward model evaluates the result. 
Prior methods make different choices within this design space. V-GPS~\citep{nakamoto2024steering} reranks single-step actions by a value model, without explicit rollout prediction. FOREWARN~\citep{wu2025foresight}, GPC~\citep{qi2025strengthening}, VLA-Reasoner~\citep{guo2025vla}, and LaDi-WM~\citep{huang2025ladi} rely on task-specific adaptation or policy refinement. In contrast, \coolname{} is a generalizable policy steering framework that performs fully deployment-time steering using a pretrained latent world model and a language-conditioned value model, without finetuning any component on target-task data. Table~\ref{tab:policy-steering-comparison} summarizes the main differences.

\begin{table}[h]
\caption{\textbf{Comparison to policy steering and model-based refinement methods.}
 \emph{Training-free composition} means fully plug-and-play, each component does not rely on other components and any target-task data. ``--'' denotes not applicable.}
\label{tab:policy-steering-comparison}
\centering
\scriptsize
\setlength{\tabcolsep}{2.5pt}
\renewcommand{\arraystretch}{0.92}

\resizebox{0.80\columnwidth}{!}{
\begin{tabular}{lccccc}

\toprule
\textbf{Method} &
\makecell{\textbf{Action}\\\textbf{chunks}} &
\makecell{\textbf{World model}\\\textbf{(space $|$ adaptation)}} &
\makecell{\textbf{Generalized}\\\textbf{evaluator}} &
\makecell{\textbf{Zero-shot}\\\textbf{steering}} &
\makecell{\textbf{Training-free}\\\textbf{composition}} \\
\midrule
V-GPS~\citep{nakamoto2024steering}
& \xmark
& --
& \cmark
& \cmark
& \xmark \\
FOREWARN~\citep{wu2025foresight}
& \cmark
& Latent $|$ task-finetuned
& \xmark
& \xmark
& \xmark \\
GPC~\citep{qi2025strengthening}
& \cmark
& Pixel $|$ task-finetuned
& \xmark
& \xmark
& \xmark \\
VLA-Reasoner~\citep{guo2025vla}
& \cmark
& Pixel $|$ generalized
& \xmark
& \xmark
& \xmark \\
LaDi-WM~\citep{huang2025ladi}
& \cmark
& Latent $|$ task-finetuned
& --
& \xmark
& \xmark \\
\midrule
\rowcolor{rowgray}
\textbf{\coolname{} (ours)}
& \cmark
& Latent $|$ generalized
& \cmark
& \cmark
& \cmark \\
\bottomrule
\end{tabular}
}
\vspace{-5pt}
\end{table}

\section{Methodology}

We consider language-instructed robot deployment in an unseen environment. At each timestep $t$, the agent receives an observation $o_t \in \mathcal{O}$ and acts according to a natural language instruction $\ell$. The action space is denoted by $\mathcal{A}$. Explicit reward signals and new demonstrations in target environment are not available. 
Our goal is to steer a pretrained generative VLA policy at test time, without any modifications to its parameters. The policy $\pi_\theta(a_{t:t+H-1} \mid o_t,\ell)$ defines a stochastic distribution over action chunks $a_{t:t+H-1} \in \mathcal{A}^{H}$ of horizon $H$. \coolname{} addresses the problem of deciding which sampled action chunk to execute by evaluating candidate futures before acting.

\begin{figure*}[t]
    \centering
    \includegraphics[width=\textwidth]{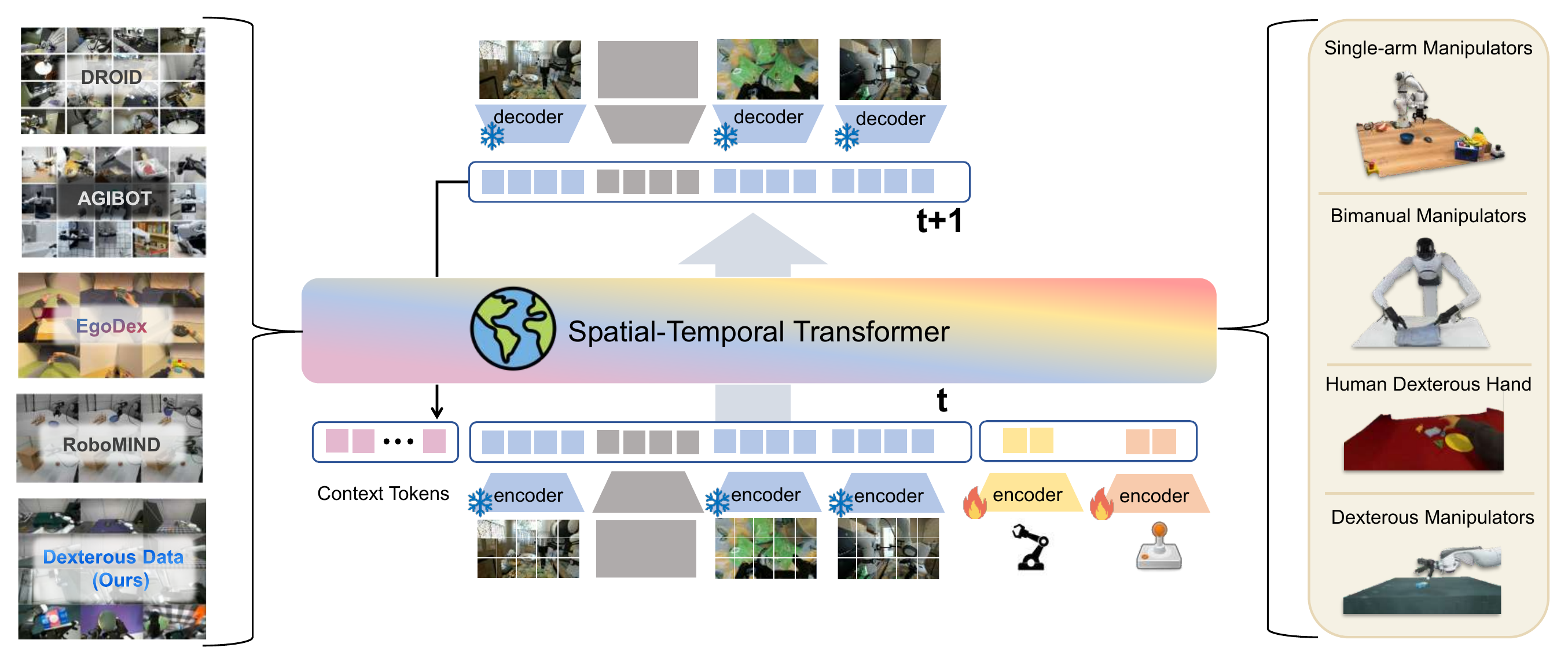}
    \caption{\textbf{Heterogeneous action-conditioned world model.}
    The world model is trained on diverse robot and human interaction datasets spanning multiple embodiments.}
    \label{fig:HWM}
\end{figure*}

\subsection{World Model Training}

The world model is a central component of \coolname{}. To support zero-shot deployment-time steering, it must be generalizable, efficient, and robust. The world model, shown in Fig.~\ref{fig:HWM}, is an action-conditioned latent dynamics model trained on heterogeneous robot and human interaction data. Its design follows three principles: multi-embodiment training for broad generalization, latent-space prediction for efficient rollout, and spatio-temporal factorization for scalable trajectory evaluation. Additional details are provided in the supplementary material Section~\ref{sec:supp-world-model}.

\textbf{Multi-Embodiment Training.}
To learn a generalizable world model, we train on heterogeneous datasets~\citep{khazatsky2024droid,bu2025agibot,hoque2025egodex,wu2024robomind}, spanning single-arm manipulators, bimanual robots, dexterous hands, and human demonstrations. Visual observations are mapped into a shared latent space, while embodiment-specific action and state inputs are encoded into latent tokens using learned tokenizers. A shared spatio-temporal transformer is trained across all embodiments. The key insight is that all these embodiments interact with a shared physical world, allowing the backbone to learn transferable object interaction dynamics despite embodiment gaps. Object interaction is both the most important and most difficult component of action-conditioned prediction: robot motion is largely specified by the action input, and static scene information is already present in the observation context, while object interaction dynamics must be learned from diverse interaction data. Missing modalities are masked during training to support partially observed multi-view and multi-embodiment data.

\textbf{Latent-Space Dynamics.}
Rather than predicting pixels directly, $\mathcal{W}_\phi$ predicts future states in the latent space of a frozen DINOv2 visual encoder. DINOv2 latent representations capture rich semantic and world knowledge while preserving task-relevant structure for downstream evaluation. Pixel-space video generation is computationally expensive and often requires iterative denoising, making it poorly suited for rollout-based steering over multiple candidates. Latent-space dynamics instead enable efficient autoregressive prediction, while decoded latent rollouts can still be evaluated by an image-based value model.
In practice, latent rollout is substantially faster than video diffusion models~\cite{guo2025ctrl}: on the same NVIDIA RTX 4090 setup, generating three $320{\times}192$ frames at horizon $H{=}10$ takes 23.12\,s with a video diffusion model, compared to 0.59\,s for our latent world model.

\begin{wrapfigure}{r}{0.6\textwidth}
    \centering
    \includegraphics[width=\linewidth]{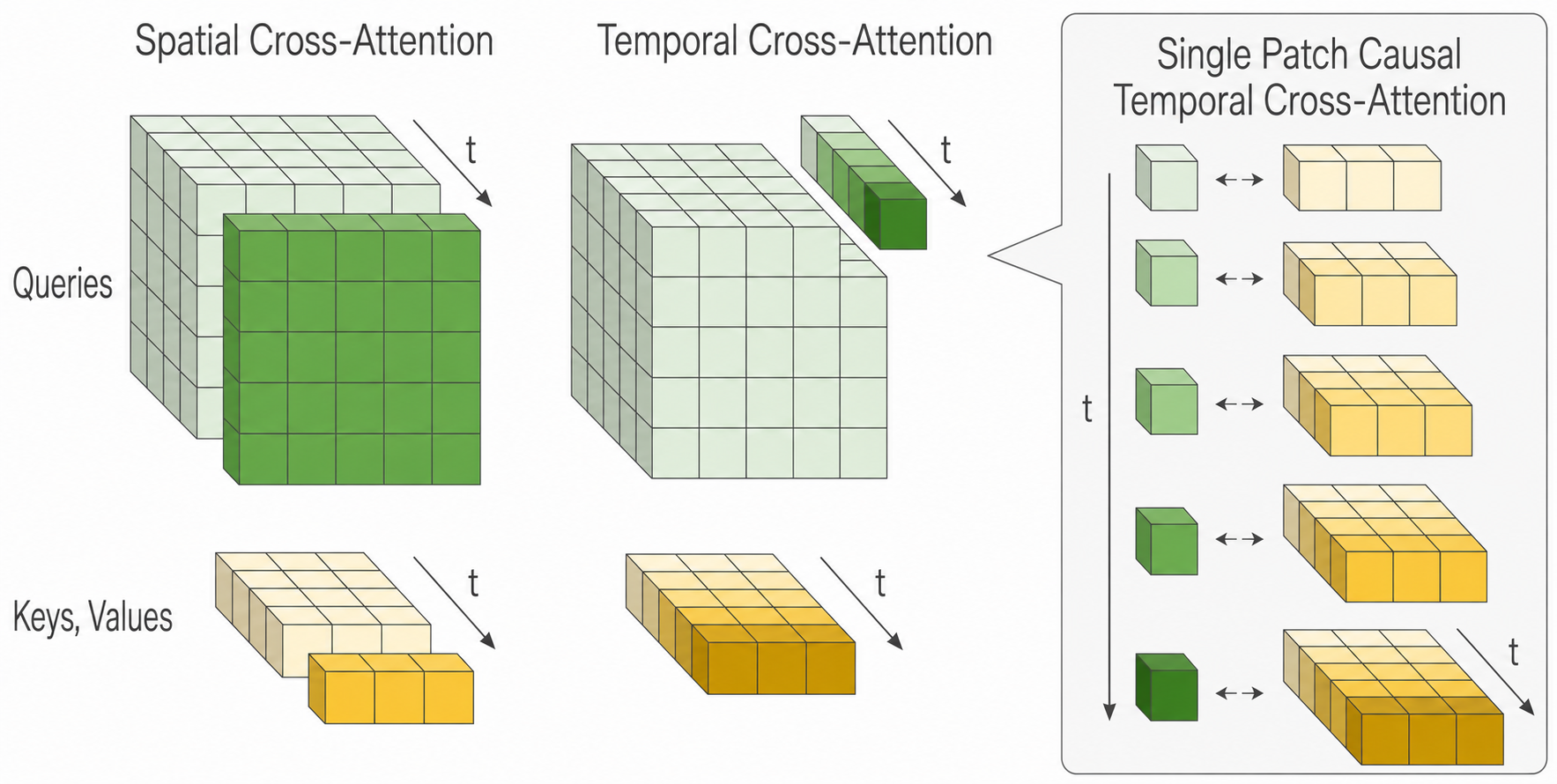}
    \caption{\textbf{Spatio-temporal cross attention.}
    Spatial cross-attention attends within each timestep, while temporal cross-attention applies causal attention across timesteps for each patch.
    }
    \label{fig:st-attention}
\end{wrapfigure}

\textbf{Spatio-Temporal Factorization.}
To support fast rollout over multiple candidate action chunks, we use a factorized spatio-temporal transformer~\citep{xu2020spatial}. The model consists of $N$ spatio-temporal blocks operating on visual latent tokens and action/state tokens. Each block applies self-attention to visual and control tokens together with cross-attention for action-conditioned prediction. As illustrated in Fig.~\ref{fig:st-attention}, attention is factorized into spatial attention within each timestep and causal temporal attention across timesteps for each patch. This factorization reduces attention complexity in the rollout horizon $H$ from quadratic to linear, enabling efficient evaluation of many candidate action chunks during deployment.

\subsection{\coolname{}}

\coolname{}, shown in Fig.~\ref{fig:dreamsteer}, steers a pretrained VLA policy by ranking imagined outcomes of candidate action chunks before execution. At time $t$, \coolname{} constructs a finite candidate set
\[
\mathcal{C}_t = \mathcal{C}^{\mathrm{VLA}}_t \cup \mathcal{C}^{\mathrm{prim}},
\]
where $\mathcal{C}^{\mathrm{VLA}}_t$ contains stochastic samples from the VLA policy and $\mathcal{C}^{\mathrm{prim}}$ contains predefined Cartesian action primitives. Each candidate action chunk is rolled out through the world model, decoded into image observations, and scored by the value model under instruction $\ell$. DreamSteer then executes the candidate with the highest predicted value score:
\begin{equation}
\begin{aligned}
k^\star
&=
\text{argmax}_{k \in \{1,\dots,|\mathcal{C}_t|\}}
V_\psi\!\left(
    o_t,
    \mathcal{W}_\phi(o_t,a^{(k)}_{t:t+H-1}),
    \ell
\right), 
&a^\star_{t:t+H-1}
= a^{(k^\star)}_{t:t+H-1}.
\end{aligned}
\end{equation}
Here, $\pi_\theta$, $\mathcal{W}_\phi$, and $V_\psi$ denote the policy, world model, and value model, respectively, and all remain fixed during deployment.

\paragraph{Stochastic Policy Action Proposals.}
The policy $\pi_\theta$ is instantiated as a pretrained VLA model. We write the per-step observation as
\[
o_t =
\{
I_t^{\mathrm{wrist}},
I_t^{\mathrm{ext}},
s_t^{\mathrm{robot}},
s_t^{\mathrm{gripper}}
\},
\]
where $I$ denotes visual observations and $s$ denotes proprioceptive state. Conditioned on $o_t$ and $\ell$, the policy samples $K_{\mathrm{VLA}}$ temporally extended action chunks:
\[
a^{(k)}_{t:t+H-1}
\sim
\pi_\theta(\cdot \mid o_t,\ell),
\qquad
k=1,\dots,K_{\mathrm{VLA}}.
\]
These samples expose diverse behaviors already present in the generative action distribution of the pretrained policy.

We augment these policy samples with a small primitive library $\mathcal{C}^{\mathrm{prim}}$. The primitive library consists of fixed short-horizon Cartesian motions, including end-effector translations along left/right, up/down, and forward/backward directions, as well as gripper open and close commands. Each primitive is represented as an action chunk of length $H$, making it temporally aligned with VLA-generated chunks. The primitives are not intended to solve manipulation tasks on their own. Instead, they provide simple alternatives that improve candidate coverage when the VLA samples alone fail to make progress, for example when the end effector is near the target object but the sampled actions do not move into a grasp-ready pose.

\paragraph{World Model Predictive Visual Dynamics.}
The world model $\mathcal{W}_\phi$ predicts future observations conditioned on the current observation and a candidate action chunk:
\[
\hat{o}^{(k)}_{t+1:t+H}
=
\mathcal{W}_\phi(o_t,a^{(k)}_{t:t+H-1}).
\]
Internally, the model operates in the latent space of a frozen DINOv2 visual encoder \citep{oquab2023dinov2}. Suppressing modality-specific tokenization for clarity, the rollout can be written as
\begin{equation}
z_t = E(o_t), \quad
\hat{z}^{(k)}_{t+1:t+H}
=
F_\phi\!\left(z_t,a^{(k)}_{t:t+H-1}\right), \quad
\hat{o}^{(k)}_{t+1:t+H}
=
D\!\left(\hat{z}^{(k)}_{t+1:t+H}\right).
\end{equation}
where $E$ is the frozen observation encoder, $F_\phi$ is the learned latent dynamics model, and $D$ is the frozen decoder that reconstructs visual observations from predicted latents. 

\paragraph{Value Function Scoring and Steering Decision.}
The value function $V_\psi$ is instantiated as a pretrained Vision-Language-Action-Critic (VLAC) model \citep{zhai2025vision} based on the InternVL2-2B architecture \citep{chen2024expanding}. VLAC provides an instruction-conditioned progress signal over pairs of visual observations. Given an imagined rollout for candidate $k$, we compute a trajectory-level score by summing pairwise progress estimates along the decoded rollout:
\begin{equation}
S^{(k)}
=
\sum_{j=1}^{H}
\mathrm{VLAC}
\!\left(
    \hat{o}^{(k)}_{t+j-1},
    \hat{o}^{(k)}_{t+j},
    \ell
\right),
\qquad
\hat{o}^{(k)}_{t} = o_t .
\end{equation}
This score estimates how much the candidate action chunk advances the task specified by $\ell$. DreamSteer does not require the value model to produce calibrated absolute rewards; it only needs the scores to rank candidates generated at the same timestep. The system then executes the action chunk with the highest score, steering the pretrained VLA policy toward predicted outcomes that better satisfy the language instruction.

\section{Experiments}

We evaluate \coolname{} on a real-world robot setup to study three questions: whether latent world-model rollouts preserve task-relevant information for value-based ranking, whether \coolname{} improves robustness under deployment-time distribution shift, and how action diversity and value-guided steering contribute to performance. Across all experiments, \coolname{} is \textbf{used zero-shot}, i.e. none of the policy, world model, or value model components are finetuned on data from the target evaluation environment.

\subsection{Real Robot Setup}

Our experiments follow the DROID setup using a 7-DoF Franka Panda manipulator with a Robotiq two-finger gripper. Compared to the original DROID configuration, we replace the exterior cameras with Intel RealSense L515 cameras and mount the robot on a static workbench.
\textbf{All evaluations are conducted in a different lab environment}, introducing a different robot instance, low-level controller, camera configuration, background, lighting, and object distribution relative to \coolname{} development. These changes occur simultaneously, resulting in substantial distribution shift.

\subsection{Evaluating the World Model}

\begin{wrapfigure}{r}{0.7\textwidth}
    \centering
    \includegraphics[width=\linewidth]{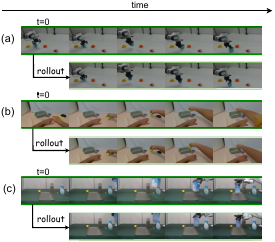}

    \caption{\textbf{World model rollouts across multiple embodiments.}
    Only a single camera view is visualized for clarity.}
    \label{fig:rollout}
\end{wrapfigure}

\begin{figure}[t]
    \centering
    \includegraphics[width=\linewidth]{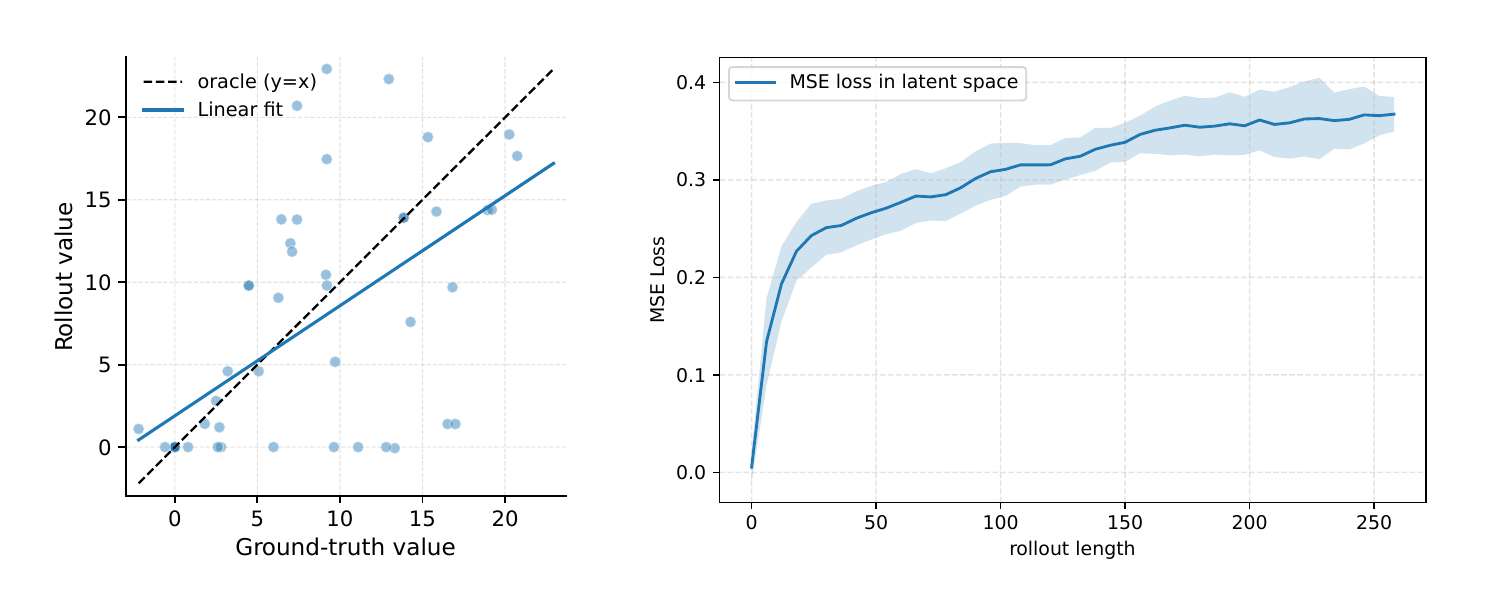}
    \caption{\textbf{Evaluative consistency of imagined rollouts.} (a) Value-model scores on ground-truth and imagined trajectories. (b) Latent prediction error versus rollout length; shaded area denotes $\pm$1 standard deviation.}
    \label{fig:VLAC_eval}
\end{figure}

We first evaluate whether latent world model rollouts preserve the relative value trends needed for trajectory ranking. \coolname{} does not require calibrated reward prediction or photorealistic video generation, only imagined rollouts that preserve relative ranking consistency across candidate action chunks. The rollout results are shown in Fig.~\ref{fig:rollout}. 
More visualizations are shown in Fig.~\ref{fig:wm_rollout_all} in supplementary material.
We compare value-model scores on short ground-truth video clips and corresponding imagined rollouts generated from action chunks with horizon $H{=}10$ under the same language instructions. As shown in Fig.~\ref{fig:rollout}, the two sets of scores are positively correlated, with Pearson $r=0.66$ (95\% CI $[0.46,0.79]$, $p<10^{-6}$) and Spearman $\rho=0.69$ ($p<10^{-7}$). This consistency is sufficient for rollout ranking in \coolname{}. We also measure latent prediction error as rollout length increases. As shown in Fig.~\ref{fig:VLAC_eval}, latent MSE increases gradually over the evaluated horizon, suggesting that the model remains sufficiently stable for repeated short-horizon rollout evaluation. All the data for quantitative evaluation are sampled from the unseen RoboArena dataset~\citep{atreya2025roboarena}.

\subsection{Evaluating \coolname{}}
\label{sec:Eval_dreamsteer}

\begin{figure}[t]
  \centering
  \includegraphics[width=\linewidth]{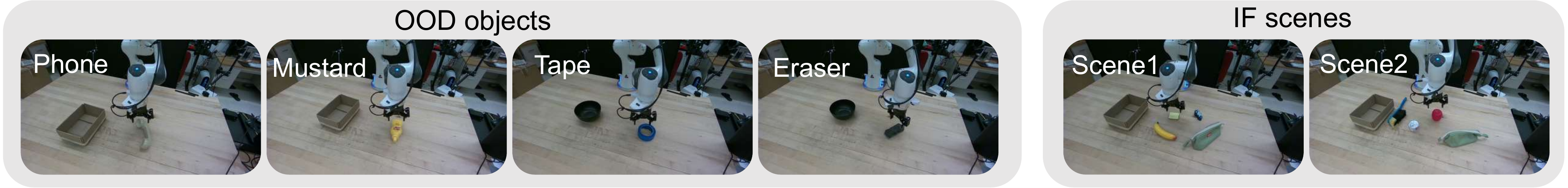}
  \caption{\textbf{Real-world evaluation tasks.} Left: OOD object manipulation tasks involving previously unseen objects. Right: instruction-following (IF) scenes containing multiple objects and distractors.}
  \label{fig:objects}
\end{figure}

We evaluate \coolname{} as a complete system with two primary goals: \textbf{(1) improving robustness to out-of-distribution (OOD) objects}, and \textbf{(2) improving instruction-following (IF) accuracy under language-specified constraints}. The OOD objects and the scenes used to evaluate IF accuracy are shown in Fig.~\ref{fig:objects}.

Across all experiments, we use a $\pi_0$ checkpoint pretrained on DROID as the base policy. Unless otherwise specified, we use $K{=}5$ sampled action chunks and horizon $H{=}10$. We use chunk-level steering because imagined observations for $H{=}1$ are often too similar for reliable ranking, while longer action chunks produce more distinguishable future states. Steering is applied once every five control steps for efficiency. Each task is evaluated over 20 trials with randomized object poses and positions. The steering step requires approximately 13\,s of computation. This cost scales with the number of candidate action chunks: generating, rolling out, and evaluating a single candidate requires approximately 1\,s in our implementation, and we use 13 candidates during deployment. Rollout and evaluation are fully parallelizable and could substantially reduce steering latency, as our current implementation prioritizes feasibility over inference optimization.

\paragraph{OOD Object Generalization.}

We evaluate \coolname{} on OOD object manipulation using a standard pick-and-place task, where the robot must place a specified object into a designated receptacle. The task structure is fixed, but the target objects differ from the policy training distribution in appearance, geometry, and material properties, making them challenging for the pretrained policy.
We compare \coolname{} against several ablations: $\pi_0$ denotes single-sample policy execution; $\pi_0$ + \coolname{} ranks multiple $\pi_0$ action chunks; primitives + \coolname{} ranks only predefined Cartesian primitives; and $\pi_0$ + primitives + random selects uniformly at random from the same candidate set as the full method.

As shown in Table~\ref{tab:ood-performance}, \coolname{} improves OOD manipulation success rate from 23.75\% to 66.25\%. Ranking multiple $\pi_0$ samples improves over single-sample execution, suggesting that useful action chunks exist in the policy distribution but are not reliably selected. Primitives alone achieve zero success, while random selection over the same candidate set also fails, indicating that both candidate diversity and value-guided ranking are necessary. The full method performs best by combining diverse policy proposals, primitive-based local exploration, and rollout evaluation.

We also study the effect of proposal count $K$ on the whiteboard eraser task. Increasing $\pi_0$ samples from $K{=}3$ to $K{=}5$ improves success, while increasing further to $K{=}7$ provides little additional benefit but increases rollout cost. We therefore use $K{=}5$ in all experiments.

\begin{table}[h]
\caption{\textbf{OOD object performance.}
Success rates over 20 trials per object. The last column reports the aggregate success rate over all 80 trials with 95\% Wilson confidence intervals.}
\label{tab:ood-performance}

\centering
\scriptsize
\setlength{\tabcolsep}{3.0pt}
\renewcommand{\arraystretch}{0.82}

\begin{tabular}{lcccccc}
\toprule
Method & Phone & Mustard & Tape & Eraser & Average & 95\% CI \\
\midrule
$\pi_0$
& 4/20 & 3/20 & 6/20 & 6/20 & 23.75 & [15.84, 34.07] \\

$\pi_0$ + \coolname{}
& 7/20 & 6/20 & 11/20 & 10/20 & 42.50 & [32.26, 53.43] \\

$\pi_0$ + primitives + random
& 0/20 & 0/20 & 0/20 & 0/20 & 0.00 & [0.00, 4.58] \\

primitives + \coolname{}
& 0/20 & 0/20 & 0/20 & 0/20 & 0.00 & [0.00, 4.58] \\

\midrule
$\pi_0$ + primitives + \coolname{}
& \textbf{12/20}
& \textbf{11/20}
& \textbf{16/20}
& \textbf{14/20}
& \textbf{66.25}
& \textbf{[55.39, 75.65]} \\
\bottomrule
\end{tabular}

\vspace{-6pt}
\end{table}

\paragraph{Instruction Following Accuracy.}

We next evaluate whether \coolname{} improves instruction following in complex scenes.  We report instruction-following accuracy. A trial is counted as correct if the robot makes consistent contact with, or attempts to grasp, the language-specified object. This metric isolates semantic target selection from low-level execution difficulty, allowing us to evaluate whether test-time steering improves adherence to language constraints. As shown in Table~\ref{tab:instruction-following}, \coolname{} improves instruction-following accuracy from 38.75\% to 56.25\%, a gain of 17.5 percentage points over $\pi_0$. These results suggest that imagined rollout scoring helps reject action proposals whose predicted outcomes do not align with the semantic intent of the instruction.

\begin{table}[h]
\caption{\textbf{Instruction following performance.}
Accuracy over 20 trials per target object. The last column reports the aggregate accuracy over all 80 trials with 95\% Wilson confidence intervals.}
\label{tab:instruction-following}

\centering
\scriptsize
\setlength{\tabcolsep}{3pt}
\renewcommand{\arraystretch}{0.82}

\begin{tabular}{lcccccc}
\toprule
Method & Sponge & Banana & Pencil & Apple & Average & 95\% CI \\
\midrule

$\pi_0$
& 8/20 & 9/20 & 6/20 & 8/20
& 38.75
& [28.78, 49.73] \\

\midrule

$\pi_0$ + primitives + \coolname{}
& \textbf{14/20}
& \textbf{13/20}
& \textbf{9/20}
& \textbf{9/20}
& \textbf{56.25}
& \textbf{[45.34, 66.57]} \\

\bottomrule
\end{tabular}

\vspace{-6pt}
\end{table}

\section{Discussion and Limitations}

\subsection{Why \coolname{} Works}

\paragraph{Trajectory Ranking Over One-pass Generation.}
\coolname{} improves deployment-time robustness by converting single-sample action generation into a ranking problem over imagined futures. A pretrained VLA policy provides a distribution over plausible action chunks, but under distribution shift a single sample may fail or violate the language instruction. \coolname{} instead samples multiple candidate chunks, predicts their outcomes with a world model, ranks the imagined rollouts, and executes the highest-scoring action chunk. The performance improvements suggest that meaningful trajectories often exist in the policy distribution but are not reliably selected during single-sample execution. Selecting among candidate trajectories is easier than generating the correct trajectory in a single pass.

\paragraph{Complementary Generalization Across Components.}
Another way to understand \coolname{} is that the policy, world model, and value model are trained under different data regimes and therefore generalize differently. Generalization often depends on both the scale and diversity of training data. A VLA policy is typically trained on successful robot demonstrations collected from limited embodiments and environments. In contrast, a world model only needs to predict future observations and can leverage broader interaction data, including successful demonstrations, failures, random play, and cross-embodiment trajectories. The value model can generalize more broadly still, since image-based progress estimation can exploit large-scale visual-language pretraining together with diverse action-free human data. \coolname{} exploits this complementarity at deployment time: even when an object is out-of-distribution for the policy, it may remain in-distribution for the world model and value model, allowing successful steering.

\subsection{Limitations}
\label{sec:limitations}

\paragraph{Failure Mode Analysis.}
\coolname{} can fail for two reasons. First, steering is limited by candidate coverage: when neither the sampled policy actions nor the predefined primitives make meaningful progress toward the instruction, the system cannot recover a successful behavior. Integrating latent-space planning or trajectory optimization could help iteratively refine candidate actions using world-model feedback. Second, steering relies on accurate trajectory ranking. We observe cases where the value model assigns higher scores to suboptimal candidates, often because different action outcomes appear visually similar from the single available camera view. In such cases, the value model cannot reliably distinguish between candidate trajectories, leading to incorrect steering decisions. Future work could improve ranking reliability through multi-view observations.

\paragraph{Latency.}
World-model rollout and value-model evaluation introduce additional inference overhead. For horizon $H{=}10$, policy inference takes 0.08\,s, world-model rollout takes 0.59\,s, and value-model evaluation takes 0.37\,s on an NVIDIA RTX~4090 GPU. The current implementation prioritizes feasibility over efficiency. Both rollout and scoring could be accelerated further through parallelized rollouts, KV caching, faster attention kernels, and other runtime enhancements.

\section{Conclusion}

We introduced \textbf{\coolname{}}, a deployment-time steering framework that improves pretrained VLA policies using latent world-model rollout evaluation, \textbf{without training or finetuning any component on target-environment data}. \coolname{} combines a pretrained VLA policy, an action-conditioned latent world model, and a language-conditioned value model to rank candidate action chunks before execution. Across real-robot evaluations, \coolname{} improves OOD object manipulation success from 23.75\% to 66.25\% and instruction-following accuracy from 38.75\% to 56.25\% over the base $\pi_0$ policy. Future work could improve both sides of this decomposition: richer candidate proposal mechanisms to increase action coverage, more efficient and robust world model rollout, and multi-view value aggregation.

\clearpage
\acknowledgments{The authors would like to thank Changhyun Choi for providing the robot platform and Mingen Li for assistance with its setup. We also thank Christopher Agia for his helpful discussions and valuable insights. Finally, we thank Adam Imdieke for his hardware support on gripper adapter.}


\bibliography{example}  

\clearpage
\section*{Supplementary Materials}

\section{World Model Design and Training}
\label{sec:supp-world-model}

The world model is an action-conditioned latent predictor optimized for efficient and robust rollout, rather than for photorealistic video generation. It is trained on heterogeneous, multi-embodiment data.

\subsection{Input Modalities and Latent Encoding}
\label{sec:supp-encoding}

The world model operates on a set of latent representations derived from multiple input modalities, consisting of visual observations, robot state, and action. All inputs are aligned temporally and encoded into a unified latent space before being processed by the spatio-temporal transformer.

\paragraph{Visual observations}
RGB observations are encoded independently at each timestep using a pretrained and frozen DINOv2 encoder. Each image is mapped to a grid of latent patch embeddings,
\[
x_t \in \mathbb{R}^{H \times W \times D_x},
\]
where $H \times W$ denotes the spatial resolution of the latent grid and $D_x$ is the per-patch embedding dimension. After projection, visual latents $x_t$ are denoted as $z_t$. When an associated decoder is available, it is used only for visualization of predicted futures and not for training.

\paragraph{Action and control signal inputs}
Robot actions and robot states are represented as per-timestep control tokens. Robot actions are expressed as end-effector delta motions in Cartesian space. These deltas are computed from observed state transitions rather than directly using low-level controller commands, allowing the model to remain agnostic to embodiment-specific control interfaces. Concretely, we treat each action/state component as a separate input key and encode it using a designated tokenizer, implemented as a lightweight embodiment-specific MLP, into latent tokens,
\[
c_{t,i} = f_i(a_{t,i}) \in \mathbb{R}^{A_i \times D_c},
\]
where $a_{t,i}$ denotes the $i$-th control component (action and, when available, state) at timestep $t$, and $f_i$ is the corresponding component-specific tokenizer. $A_i$ denotes the number of tokens produced by component $i$. The per-component tokens are then concatenated along the token dimension to form the control token sequence
\[
c_t = \big[ c_{t,1}, \ldots, c_{t,K} \big] \in \mathbb{R}^{A \times D_c}, \quad A=\sum_{i=1}^{K} A_i,
\]
with $K$ components provided by the embodiment.

\paragraph{Temporal alignment and batching}
Visual and control tokens are aligned at the timestep level. Given a sequence of length $T$, the model receives visual latents
\[
x \in \mathbb{R}^{B \times T \times S \times D_x}
\]
and corresponding control tokens
\[
c \in \mathbb{R}^{B \times T \times A \times D_c},
\]
where $S$ is the total number of visual tokens per timestep, aggregated across available views, and $A$ is the number of control tokens per timestep, which may vary across embodiments. These representations are inputs to the spatio-temporal transformer, which predicts future visual latents conditioned on past observations and the provided action/state sequence.

\paragraph{Projection to model dimension}
Before entering the transformer, both visual and control embeddings are linearly projected to a shared model dimension $D$. This allows the subsequent attention layers to operate in a unified embedding space while preserving the distinct structural roles of visual grid tokens and control tokens.

\subsection{Spatio-Temporal Transformer Architecture}
\label{sec:supp-st-arch}

Given the encoded visual and control tokens, the world model predicts future visual latents using a stack of spatio-temporal transformer layers. The transformer operates on two token streams: a grid of visual tokens and a set of per-timestep control tokens, and updates the visual stream through action-conditioned cross-attention.

\paragraph{Spatio-temporal transformer layers}

\begin{figure}[t]
  \centering
  \includegraphics[width=\linewidth]{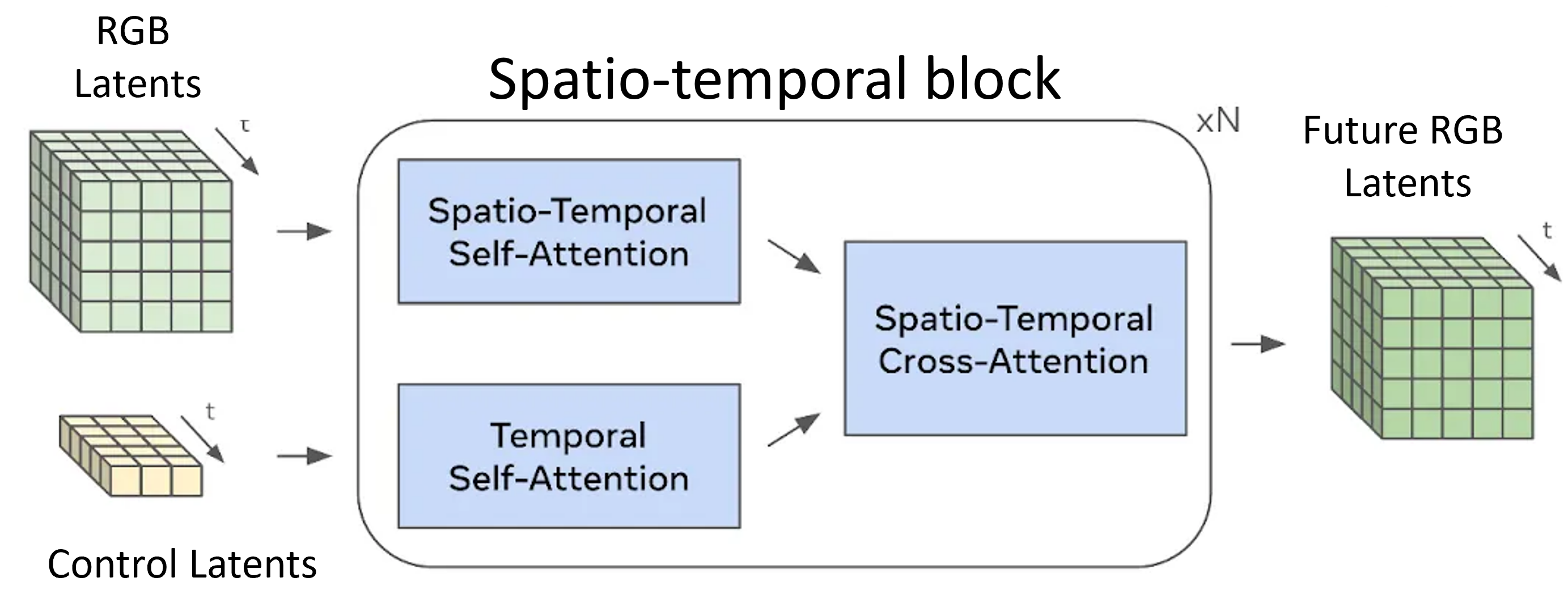}
  \caption{\textbf{Spatio-temporal world model architecture.}}
  \label{fig:st_block}
\end{figure}

The transformer consists of $N$ repeated layers, each composed of three sequential blocks, as shown in Fig.~\ref{fig:st_block}: (i) spatio-temporal self-attention over visual tokens, (ii) temporal self-attention over control tokens, and (iii) spatio-temporal cross-attention from visual tokens to control tokens. Spatio-temporal self-attention is factorized into spatial attention applied independently within each timestep and causal temporal attention applied across timesteps at each spatial (or control) location. This factorization avoids full attention over all space-time tokens and improves computational efficiency for long-horizon rollouts. Action conditioning is introduced through cross-attention, where visual tokens act as queries and control tokens serve as keys and values in a time-aligned manner, which is shown in Fig.~\ref{fig:st-attention}.

\subsection{Train with Multi-Embodiment Data}
\label{sec:supp-mult-embodiment}

The world model is trained on data collected from multiple robot embodiments with heterogeneous sensory configurations and control interfaces. Differences across embodiments are handled at the input tokenization level, while all subsequent spatio-temporal dynamics parameters are fully shared.

\paragraph{Training datasets}
The world model is trained on a mixture of multi-embodiment manipulation datasets spanning both robot and human demonstrations. Our training corpus includes: (i) the DROID dataset collected on Franka platforms (76k trajectories, $\sim$350 hours), (ii) Franka subsets from RoboMIND, (iii) the EgoDex dataset consisting of human dexterous hand interactions, (iv) the AgiBot dual-arm manipulation dataset, and (v) an in-house teleoperation dataset containing approximately 1.2k trajectories collected on Franka arms equipped with dexterous hands.

In total, the combined dataset comprises on the order of $10^5$ trajectories and several hundred hours of interaction data. The data spans single-arm and dual-arm robots, parallel grippers and dexterous hands, as well as human hand demonstrations, covering diverse viewpoints and control interfaces. This heterogeneity enables the world model to learn embodiment-agnostic interaction dynamics while remaining compatible with varied sensory and action configurations.

\paragraph{Multi-view visual observations}

\begin{figure}[t]
  \centering
  \includegraphics[width=\linewidth]{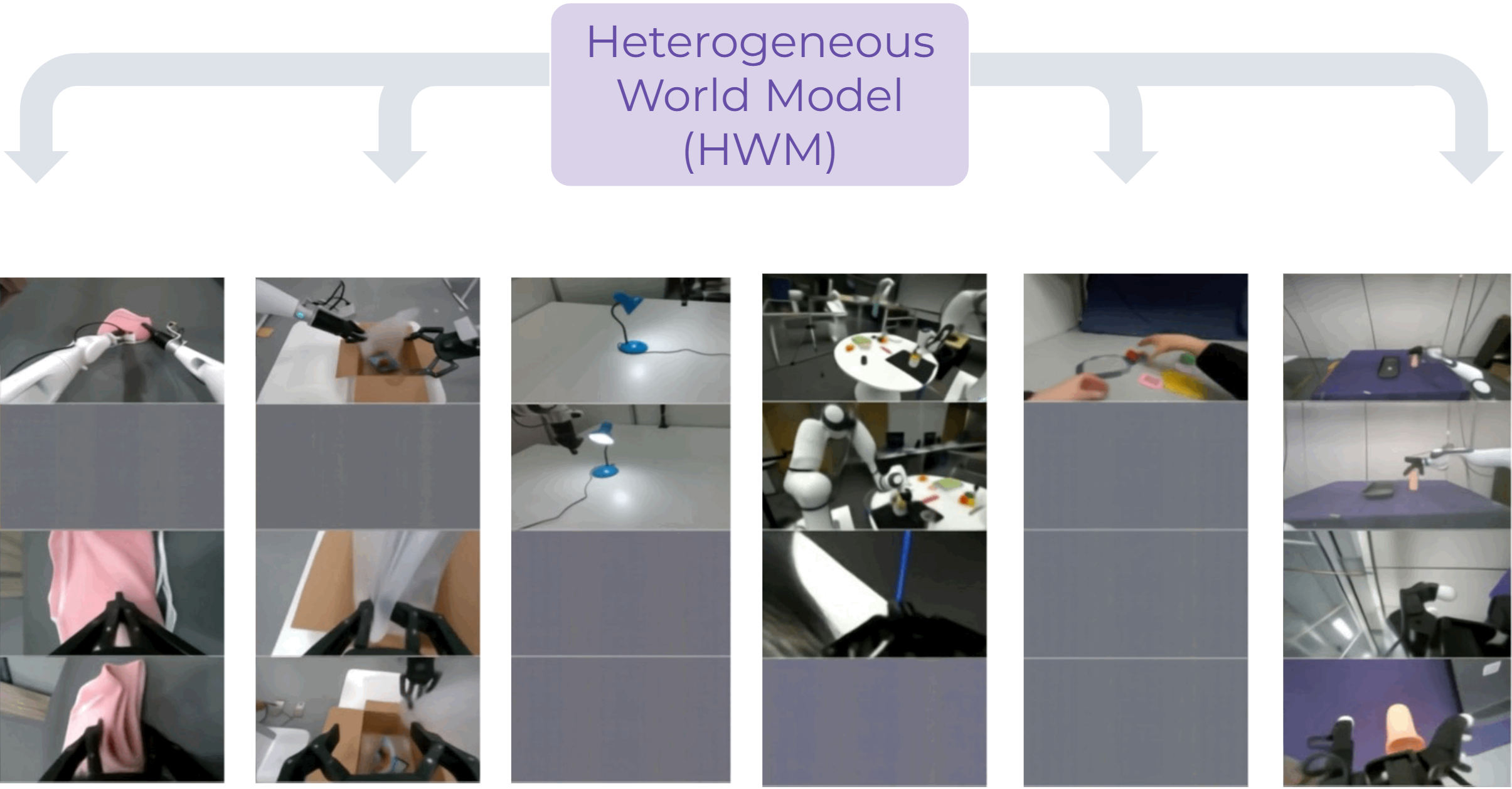}
  \caption{\textbf{Mask strategy in multi-view data.}}
  \label{fig:multimodal_mask}
\end{figure}

Each timestep may include up to four RGB observations, consisting of two external views and two wrist-mounted views. Let
\[
\mathcal{V} = \{v_1, \dots, v_{|\mathcal{V}|}\}, \quad |\mathcal{V}| \leq 4
\]
denote the set of available viewpoints at a given timestep. Each view $v \in \mathcal{V}$ is independently encoded by a pretrained visual tokenizer into a grid of latent tokens. Since not all viewpoints are available for every timestep or embodiment, we employ token-level masking to handle missing observations, shown in Fig.~\ref{fig:multimodal_mask}. For each view $v$, a binary mask
\[
m_t^{(v)} \in \{0,1\}
\]
is expanded to all corresponding latent tokens and applied prior to the dynamics model. Masked tokens are zeroed out and excluded from loss computation, enabling the model to operate on partially observed multi-view inputs without introducing padded or hallucinated observations.

\paragraph{Shared dynamics across embodiments}
After tokenization and positional embedding, implemented via rotary position embeddings (RoPE), both visual tokens $x_t$ and control tokens $c_t^{(e)}$ are projected to a shared model dimension and processed by a spatio-temporal transformer. All transformer parameters are shared across embodiments; the only embodiment-specific components are the input tokenizers for actions and states. As a result, the model learns a unified latent dynamics representation while remaining compatible with heterogeneous sensory layouts and control interfaces.

\subsection{Autoregressive Rollout and Training Objective}
\label{sec:supp-rollout-loss}

\paragraph{Problem formulation}
Let $\mathbf{o}_{1:T}$ denote a sequence of observations and $\mathbf{a}_{1:T}$ the corresponding action/state tokens. After modality-specific tokenization (Sec.~\ref{sec:supp-encoding}), observations are represented as latent tokens
\[
\mathbf{z}_t \in \mathbb{R}^{S \times D},
\]
where $S$ is the number of visual tokens (aggregated across views) and $D$ is the shared model dimension. Actions and proprioceptive states are encoded as control tokens
\[
\mathbf{c}_t \in \mathbb{R}^{A \times D}.
\]

The world model learns a dynamics function
\[
f_\theta:\; (\mathbf{z}_{1:t}, \mathbf{c}_{1:t}) \;\rightarrow\; \hat{\mathbf{z}}_{t+1},
\]
implemented by the spatio-temporal transformer described in Sec.~\ref{sec:supp-st-arch}.

\paragraph{Sliding-window autoregressive rollout}
At both training and inference time, future observations are predicted autoregressively. Given an initial context of length $T_c$, predictions are generated sequentially:
\[
\hat{\mathbf{z}}_{t+1} = f_\theta(\mathbf{z}_{t-T_c+1:t},\; \mathbf{c}_{t-T_c+1:t}),
\]
where only the most recent $T_c$ steps are retained to bound memory and computation. Predicted tokens are appended to the context and used to predict subsequent steps:
\[
\hat{\mathbf{z}}_{t+k} = f_\theta(\hat{\mathbf{z}}_{t+k-T_c:t+k-1},\; \mathbf{c}_{t+k-T_c:t+k-1}).
\]

This sliding-window mechanism enables long-horizon rollout while maintaining fixed computational cost per step.

\paragraph{Teacher-forcing objective}
For one-step prediction, the model is trained using teacher forcing, where ground-truth observations are provided as input context. The loss is computed as mean squared error (MSE) in latent space:
\[
\mathcal{L}_{\text{1-step}}
= \frac{1}{T-1} \sum_{t=1}^{T-1}
\left\| \hat{\mathbf{z}}_{t+1} - \mathbf{z}_{t+1} \right\|_2^2.
\]

When modality masks are present (e.g., missing views), the loss is computed only over valid tokens:
\[
\mathcal{L}_{\text{1-step}}
= \frac{\sum m_{t,s}\,
\|\hat{\mathbf{z}}_{t+1,s} - \mathbf{z}_{t+1,s}\|_2^2}
{\sum m_{t,s}},
\]
where $m_{t,s}\in\{0,1\}$ denotes token validity.

\paragraph{Open-loop sampling objective}
To improve long-horizon stability, we additionally train the model using open-loop rollout. Starting from a short ground-truth context (typically one frame), the model generates predictions autoregressively for a rollout horizon $H_{\text{rollout}}$:
\[
\hat{\mathbf{z}}_{2:H_{\text{rollout}}+1}
=
\text{Rollout}_\theta
\!\left(
\mathbf{z}_1,\;
\mathbf{c}_{1:H_{\text{rollout}}}
\right).
\]

A final forward pass is then performed using the generated trajectory as context, and predictions are supervised against ground truth:
\[
\mathcal{L}_{\text{n-step}}
=
\frac{1}{H_{\text{rollout}}}
\sum_{t=1}^{H_{\text{rollout}}}
\left\|
\hat{\mathbf{z}}_{t+1}
-
\mathbf{z}_{t+1}
\right\|_2^2 .
\]

This objective encourages the model to remain stable under its own predictions and mitigates exposure bias.

\paragraph{Combined training objective}
During training, we alternate between teacher-forcing and sampling losses using a scheduled curriculum. At each optimization step, only one objective is active:
\[
\mathcal{L} =
\begin{cases}
\mathcal{L}_{\text{1-step}}, & \text{teacher-forcing step} \\
\mathcal{L}_{\text{n-step}}, & \text{sampling step}.
\end{cases}
\]

Both objectives are computed as mean squared error (MSE) in the latent space of the frozen visual encoder:
\[
\mathcal{L}
= \mathbb{E}\big[
\|\hat{\mathbf{z}} - \mathbf{z}\|_2^2
\big].
\]

No pixel-space reconstruction loss is used. Supervising predictions in latent space significantly reduces computational cost while preserving task-relevant dynamics required for rollout evaluation.

\paragraph{Training details and hyperparameters}
We train the world model using distributed data-parallel training on 384 NVIDIA H100 GPUs for approximately 2-3 days. Training follows the objectives described in Sec.~\ref{sec:supp-rollout-loss}, with frozen visual tokenizers and jointly optimized action/state tokenizers. All hyperparameters are listed in Table~\ref{tab:training-hparams}.

\paragraph{Inference}
At deployment time, the world model operates purely autoregressively. Given past observations and candidate action sequences, future latent trajectories are rolled out and used for downstream evaluation without access to ground-truth future observations.

\begin{table}[h]
\centering
\small
\begin{tabular}{l l}
\toprule
\textbf{Hyperparameter} & \textbf{Value} \\
\midrule

\multicolumn{2}{l}{\textit{World Model Architecture}} \\
Model dimension $D$ & 1536 \\
Transformer layers & 8 \\
Attention heads & 24 \\
Visual latent dim $D_x$ & 1024 \\
Action/state latent dim & 16 \\

\midrule
\multicolumn{2}{l}{\textit{Context and Rollout}} \\
Training context length $T_c$ & 16 \\
Rollout horizon $H_{\text{rollout}}$ & 4 \\

\midrule
\multicolumn{2}{l}{\textit{Optimization}} \\
Optimizer & AdamW \\
Learning rate & $1\times10^{-4}$ \\
Weight decay & $1\times10^{-2}$ \\
Gradient clipping & 1.0 \\
Batch size  & 384 \\
LR schedule & Cosine decay \\

\midrule
\multicolumn{2}{l}{\textit{Loss}} \\
Latent loss type & MSE (L2) \\
Teacher forcing & Yes \\
Sampling loss & Yes \\
Loss scheduling & Alternating curriculum \\

\bottomrule
\end{tabular}
\caption{Training hyperparameters for the spatio-temporal world model.}
\label{tab:training-hparams}
\end{table}

\subsection{World Model Rollout Visualization}
\label{sec:supp-rollout-vis}

\begin{figure}[t]
  \centering
  \includegraphics[width=\linewidth]{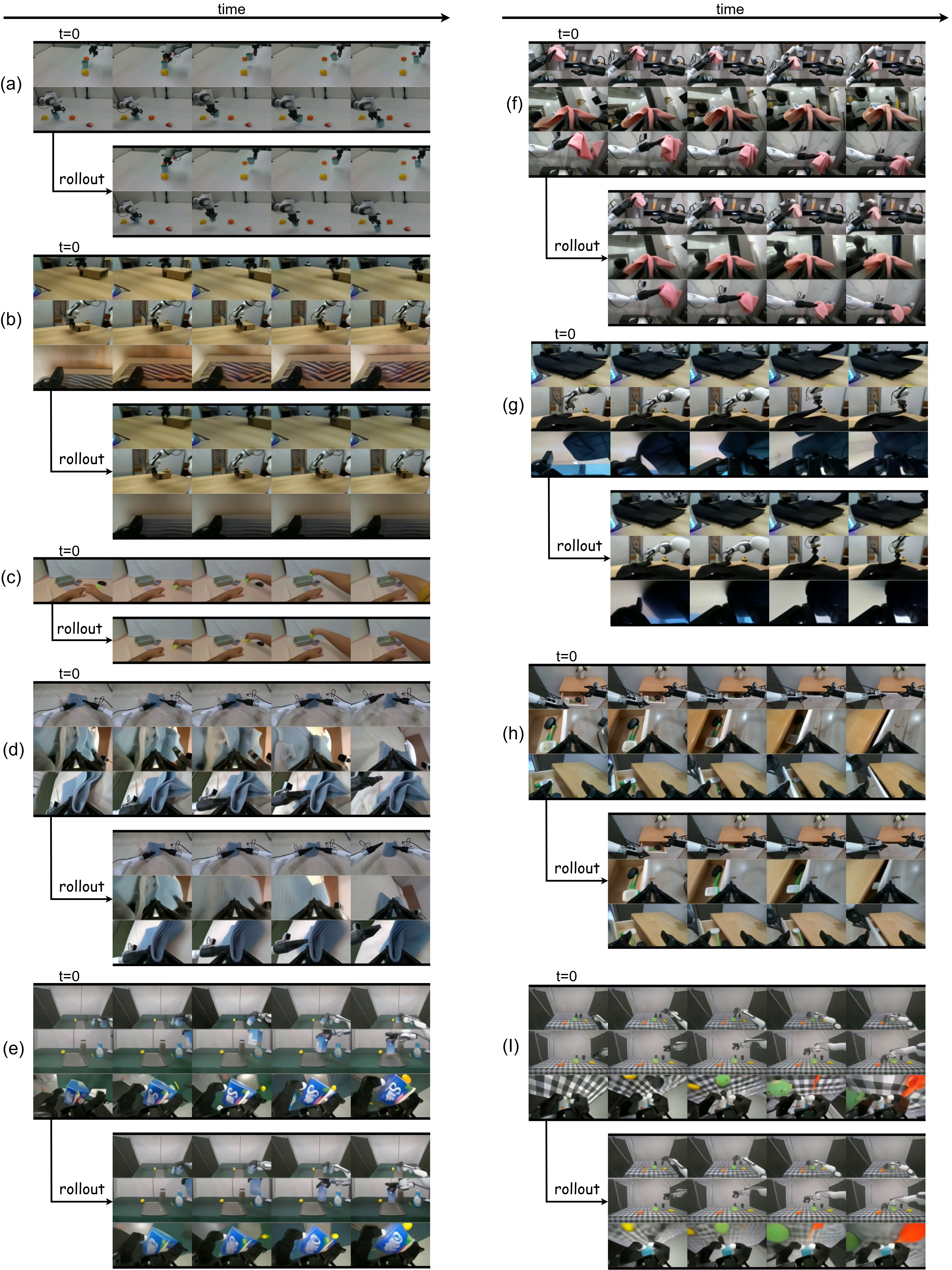}
  \caption{\textbf{World model rollout results.}}
  \label{fig:wm_rollout_all}
\end{figure}

We provide qualitative visualizations of autoregressive world model rollouts across multiple embodiments, viewpoints, and manipulation scenarios. In all visualizations, the model predicts future visual latents autoregressively conditioned on past observations and the provided action sequence. The rollout visualizations are shown in Fig.~\ref{fig:wm_rollout_all}.

Unless otherwise specified, rollouts are initialized from a single ground-truth observation frame, after which predictions are generated fully open-loop. We visualize decoded RGB observations obtained from the frozen visual tokenizer decoder for interpretability. Ground-truth and predicted frames are shown side-by-side for comparison.

\begin{figure}[t]
  \centering
  \includegraphics[width=0.8\linewidth]{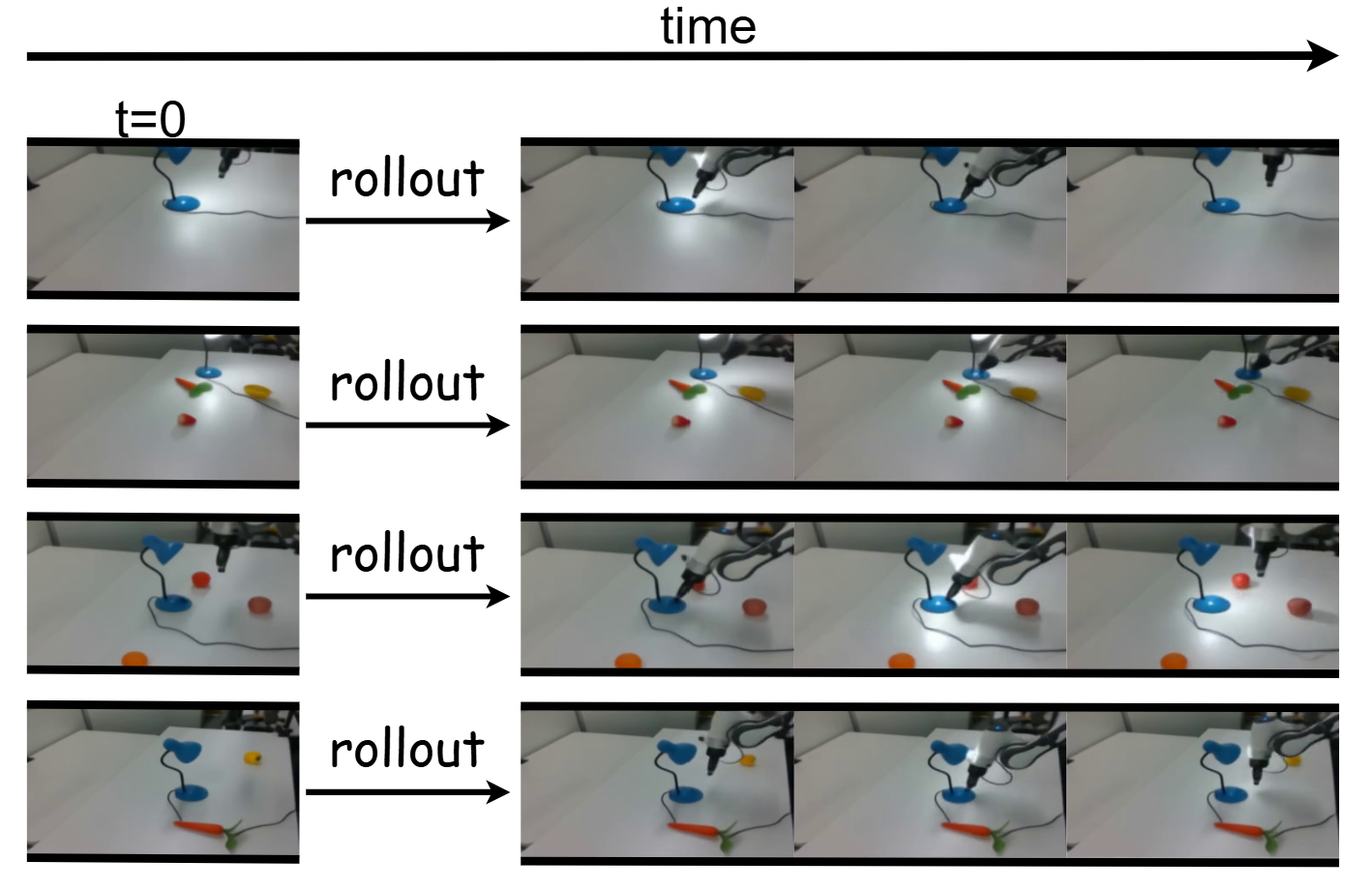}
  \caption{\textbf{Functional interaction prediction.} When the robot toggles the light switch in imagination, the world model predicts corresponding changes in scene illumination.}
  \label{fig:light}
\end{figure}

Beyond low-level physical motion prediction, we examine whether the world model captures higher-level regularities of object functionality and human-designed interactions. As shown in Fig.~\ref{fig:light}, we consider imagined rollouts in which a robot interacts with a light switch. Conditioned on the toggling action, the model predicts a corresponding change in the environmental state, such as the light turning on or off. This behavior extends beyond immediate contact dynamics and reflects learned associations between object affordances and their functional effects. Such results suggest that the world model internalizes commonsense interaction patterns present in human environments, enabling it to anticipate functional outcomes of actions in addition to physical motion.

\section{Value Model and Trajectory Scoring}
\label{sec:supp-value}

\paragraph{Model choice}
To evaluate predicted rollouts during deployment-time steering, we adopt an off-the-shelf vision--language value model, Vision-Language-Action-Critic (VLAC), without any additional finetuning. VLAC is pretrained to estimate task progress between pairs of observations conditioned on a language instruction, producing a scalar progress score that reflects relative advancement toward task completion.

\paragraph{Rollout scoring formulation}
Given the current observation $o_t$ and a candidate action sequence $a_{t:t+H-1}$, the world model predicts a rollout of future observations
\[
\hat{o}_{t+1:t+H}.
\]

For a candidate rollout, we compute pairwise progress estimates between consecutive observations:
\[
s_{t+j}
=
\mathrm{VLAC}
\!\left(
\hat{o}_{t+j-1},
\hat{o}_{t+j},
l_{\mathrm{task}}
\right),
\qquad
\hat{o}_t = o_t ,
\]
where $l_{\mathrm{task}}$ denotes the language instruction.

\paragraph{Trajectory aggregation}
The trajectory-level score is obtained by summing progress estimates along the rollout:
\[
S
=
\sum_{j=1}^{H}
\mathrm{VLAC}
\!\left(
\hat{o}_{t+j-1},
\hat{o}_{t+j},
l_{\mathrm{task}}
\right).
\]

This score estimates how much the candidate action sequence advances the task specified by the language instruction. During deployment-time steering, multiple candidate action sequences are evaluated and ranked according to their trajectory scores, and the highest-scoring candidate is selected for execution. The value model operates on a single left-view RGB observation.

\section{Action Candidate Construction}
\label{sec:supp-candidates}

The pretrained $\pi_{0}$ policy outputs fixed-length action chunks in joint space with horizon $T=10$:
\[
\mathbf{q}_{t:t+T-1}.
\]
We convert these joint actions into end-effector Cartesian delta actions using forward kinematics (FK):
\[
\Delta \mathbf{p}_t = \mathrm{FK}(\mathbf{q}_{t+1}) - \mathrm{FK}(\mathbf{q}_t),
\]
resulting in Cartesian action chunks of the same length $T=10$.

In addition to policy-generated actions, we construct a set of hand-designed Cartesian action primitives, including directional motions (up, down, forward, backward, left, right) and gripper commands (open, close).

Each primitive defines a fixed Cartesian displacement (or gripper command), which is evenly distributed across the $T=10$ steps to form a temporally aligned action chunk.

\section{Hardware Setup and Evaluation Tasks}
\label{sec:supp-hardware}

\begin{figure}[t]
  \centering
  \includegraphics[width=0.9\linewidth]{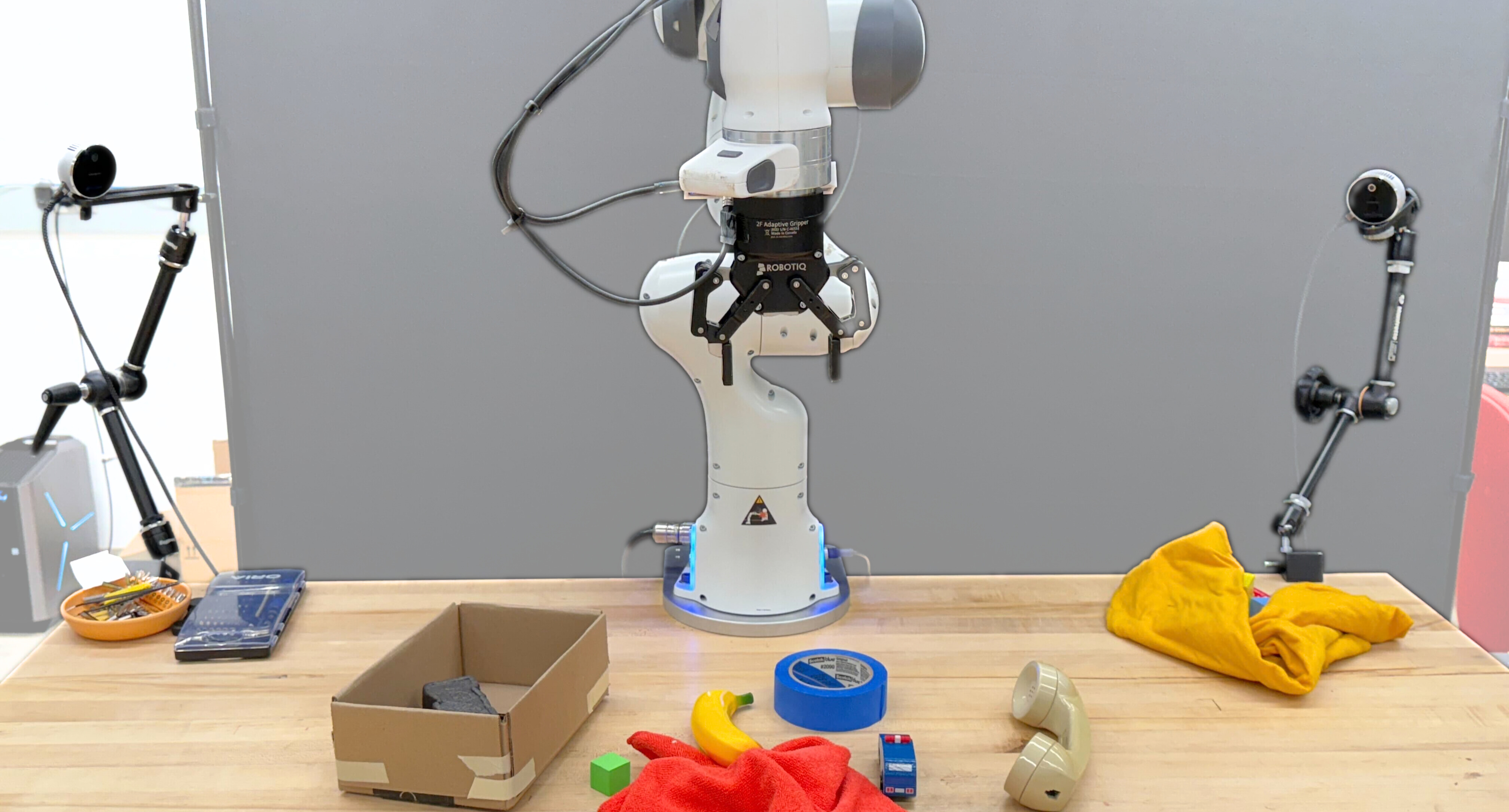}
  \caption{\textbf{Robot platform.}}
  \label{fig:robot_edit}
\end{figure}

 A visualization of the robot platform and camera configuration is shown in Fig.~\ref{fig:robot_edit}.

We evaluate real-world performance across two task families designed to assess generalization and language grounding under deployment conditions:

\begin{itemize}
    \item \textbf{Out-of-distribution (OOD) manipulation.} 
    The robot is instructed to manipulate novel objects not seen during training. Language instructions are:
    \begin{itemize}
        \item ``Pick up the \{phone\} and place it into the brown box.''
        \item ``Pick up the \{mustard\} and place it into the brown box.''
        \item ``Pick up the \{whiteboard eraser\} and place it into the black bowl.''
        \item ``Pick up the \{blue tape\} and place it into the black bowl.''
    \end{itemize}

    \item \textbf{Instruction following with distractors.} 
    The robot must identify the correct target object in the presence of visually similar distractors. Instructions and distractors are:
    \begin{itemize}
        \item ``Pick up the \{sponge\} and place it into the black bowl.''. The distractors are banana, police car toy, and pencil case.
        \item ``Pick up the \{banana\} and place it into the black bowl.'' The distractors are sponge, police car toy, and pencil case.
        \item ``Pick up the \{pencil case\} and place it into the brown box.'' The distractors are brush, can, and apple.
        \item ``Pick up the \{apple\} and place it into the brown box.'' The distractors are brush, can, and pencil case.
    \end{itemize}
\end{itemize}

\end{document}